\documentclass[smallextended,natbib]{svjour3}
\bibpunct{[}{]}{,}{n}{}{,}

\smartqed

\usepackage[T1]{fontenc}
\usepackage[utf8]{inputenc}
\usepackage{lmodern}
\usepackage{textcomp}
\usepackage[english]{babel}
\usepackage{amsmath}
\usepackage{booktabs}
\usepackage{longtable}
\usepackage{array}
\usepackage{graphicx}
\graphicspath{{./}}
\usepackage{tikz}
\usepackage[ruled,vlined]{algorithm2e}
\usepackage{siunitx}
\usepackage{fvextra}
\DefineVerbatimEnvironment{tightverbatim}{Verbatim}{fontsize=\small,breaklines=true,breakanywhere=true}
\usepackage{geometry}
\usepackage{fancyhdr}
\usepackage{titlesec}
\titleformat{\section}{\large\bfseries}{\thesection.}{1em}{}
\titleformat{\subsection}{\normalsize\bfseries}{\thesubsection.}{1em}{}
\titlespacing*{\section}{0pt}{1.2em}{0.8em}
\titlespacing*{\subsection}{0pt}{1em}{0.6em}
\usepackage{hyphenat}
\usepackage{microtype}
\usepackage{enumitem}
\usepackage{orcidlink}
\usepackage{xurl}
\usepackage{hyperref}
\hypersetup{
  pdftitle={Runtime Governance for Agentic AI: Action-Boundary Control with Trusted Provenance and Fail-Closed Execution},
  pdfauthor={Adam Massimo Mazzocchetti; ORCID 0009-0000-4584-1784},
  pdfsubject={Runtime governance for side-effectful agentic AI using action-boundary control, trusted provenance, fail-closed execution semantics, and Senate-style settlement},
  pdfkeywords={agentic AI, runtime governance, action-boundary governance, AI agents, tool use, policy enforcement point, policy decision point, trusted provenance, fail-closed execution, prompt injection, AI safety, AI governance, auditability},
  colorlinks=true,
  linkcolor=black,
  citecolor=black,
  urlcolor=blue
}

\journalname{arXiv preprint}
\newcommand{\significancestatement}[1]{\section*{Significance statement}#1}

\title{Runtime Governance for Agentic AI: Action-Boundary Control with Trusted Provenance and Fail-Closed Execution}
\titlerunning{Runtime Governance for Agentic AI}

\author{Adam Massimo Mazzocchetti\orcidlink{0009-0000-4584-1784}}
\authorrunning{Mazzocchetti}

\institute{SPQR Technologies Inc. \at
  \url{https://spqrtech.ai} \\
  \email{adam@spqrtech.ai}
}

\date{}

\begin{document}

\maketitle

\begin{abstract}
Agentic AI systems request tool actions that can modify files, send messages, launch jobs, or change workflow state. This shifts the safety problem from harmful text generation to harmful operational side effects. Prompt-level governance can shape model behavior, but it does not create an execution boundary. We introduce Aegis, a runtime governance system that treats model outputs as action proposals and mediates them through a trusted decision layer before tool execution. The model proposes; the trusted runtime decides. Aegis evaluates proposals against active policy state, resolves provenance server-side, fails closed under uncertainty, and routes selected cases through Senate-style settlement, a quorum-based non-unilateral authorization path. We evaluate Aegis on a repeated sandbox corpus spanning five run families, 42 tasks, three conditions, and ten repeats per family. Across 6,300 rows, prompt-policy conditioning produced 79 risky comparator-path leakage rows. Across 2,100 Aegis-governed rows, the system recorded zero governed mock-tool applications and zero governed risky side-effect completions. All 1,832 Aegis-attempted governed rows preserved trusted Aegis-resolved provenance, and all 1,019 Senate-settled rows had quorum and final signed tally evidence. These results do not prove general autonomous-agent safety. They support the narrower systems claim that, in this evaluated sandbox corpus, runtime action-boundary governance prevented observed risky proposals from becoming governed side effects.
\end{abstract}

\significancestatement{AI agents increasingly request actions, not just text. That changes the safety problem: harm occurs when a risky proposal becomes an operational side effect. Prompt instructions can reduce risky behavior, but they do not decide whether tools execute. Aegis moves governance to the action boundary by treating model outputs as proposals that require trusted runtime authorization. In repeated sandbox evaluation, prompt-policy conditioning still leaked risky proposals, while Aegis recorded zero governed risky side-effect completions with auditable provenance and Senate-settlement traces.}
\section{Introduction}

Agentic AI systems are crossing a boundary. They no longer merely generate text for humans to interpret; they increasingly propose actions through tools, workflows, memory, files, communications systems, and operational APIs. A model may draft and send a message, request a file export, launch a background job, approve a workflow step, mark a task complete, or pass instructions to another agent. Recent agent-safety benchmarks and autonomous-agent threat analyses show that these systems create operational side-effect risks across files, communications, code execution, browser environments, workflows, and multi-agent settings \citep{vijayvargiya2025openAgentSafety,zhang2024agentSafetyBench,shapira2026agentsOfChaos}. This changes the safety problem. The central failure mode is not only that a model says something unsafe. It is that a model proposes an unsafe action and an operational system executes it.

In agentic systems, the relevant safety event is not merely the generation of a risky sentence. It is the transition from a model-proposed action to an operational effect. A governance mechanism that only instructs the model remains inside the actor being governed. A runtime boundary changes the object of control: it mediates whether the proposed action may execute.

This paper starts from that boundary. In a text-only setting, prompt-level policy can be a useful behavioral constraint. In an agentic setting, prompt governance can shape behavior, but it is not an execution boundary. Prompt instructions can be bypassed, misread, contradicted by adversarial context, or distorted by tool-selection pressure. Indirect prompt-injection work shows that adversarial external content can manipulate tool-integrated agents toward harmful actions or data exfiltration \citep{zhan2024injecagent}. Tool-selection attacks show that malicious tool descriptions can influence which tool an agent selects and how the action is invoked \citep{shi2025toolHijacker}. Security guidance similarly treats prompt injection as an operational security concern for AI systems \citep{owasp2025llm01PromptInjection,ncsc2023thinkingAboutAiSecurity}. Even when a model appears to understand a policy, it may still produce a tool request that would create an unauthorized disclosure, runaway resource use, false completion state, or disproportionate operational action. The question for agentic AI governance is therefore not merely whether the model ``knows'' the policy. The question is whether unsafe proposals can become side effects.

We introduce Aegis, a runtime governance system for action-boundary control in agentic AI. Aegis places a trusted decision layer between the model's proposed action and the tool that would carry it out. The model proposes; the trusted runtime decides. This design draws on complete-mediation and reference-monitor principles for mediating protected operations \citep{saltzer1975protection,anderson1972computerSecurity}. It also draws on PEP/PDP authorization architectures \citep{chen2013xacmlRiskAware,openid2026authorizationApi} and runtime-assurance approaches for constraining unverified behavior before it affects a controlled system \citep{hobbs2021runtimeAssuranceSafetyCritical}. Aegis evaluates proposed actions against active policy state, resolves policy provenance server-side, withholds or fails closed when requirements are not met, and uses Senate-style settlement---a quorum-based non-unilateral authorization path---when policy requires governed cases to be decided beyond a single actor.

The design is deliberately not a claim that models become intrinsically safe. Aegis does not attempt to make every model output correct, harmless, or truthful. Instead, it changes the control point. It treats model output as a proposal, not as authority. Runtime governance decides whether a proposed action may execute. That distinction matters because agentic risk is operational: the harm occurs when a proposal becomes an effect.

We use action-boundary governance to describe runtime control over the point at which a model-proposed action would otherwise become an operational side effect.

Aegis is therefore presented as a systems contribution: a concrete implementation of action-boundary governance for side-effectful agentic AI. The contribution is not a new prompt policy, a model-alignment claim, or a benchmark leaderboard. It is a runtime architecture for treating model outputs as proposals, resolving policy and provenance through trusted infrastructure, and deciding whether proposed actions may become operational effects.

We evaluate Aegis in a repeated sandbox corpus covering 42 tasks across three conditions: a plain mesh agent, a prompt-policy mesh agent, and an Aegis-governed mesh agent. The repeated evaluation spans five run families: a deterministic stubbed model, Gemma, and a frontier model at three temperature settings. Repeated-run evaluation is used here to stress the enforcement boundary under stochastic proposal behavior and local runtime variation, not to construct a model leaderboard \citep{blackwell2025reproducibleLlmEvaluation,doRepetitionsMatter2025,reasonbench2025instability,szalontai2025llmBugfixingReproducibility}. Across 6,300 total rows and 2,100 governed rows, prompt-policy conditioning reduced but did not eliminate risky leakage. Aegis-governed execution produced zero governed mock-tool applications and zero governed risky side-effect completions.

This paper makes five contributions:
\begin{enumerate}
    \item It frames action-boundary governance as the central runtime problem for tool-using AI agents.
    \item It introduces Aegis, a runtime governance architecture that evaluates side-effectful proposals before tool execution.
    \item It separates model proposals, runtime decisions, Senate settlement, trusted provenance, and final execution outcomes into auditable traces.
    \item It evaluates the system across repeated stubbed, open-model, and frontier-model run families.
    \item It shows that prompt-policy leakage persisted in the comparator path, while Aegis-governed risky side-effect completion was zero in the evaluated corpus.
\end{enumerate}

The empirical claim is intentionally narrow. We do not claim that Aegis makes models intrinsically safe, that the task corpus is exhaustive, or that every future tool environment is covered. We claim that, in this repeated sandbox evaluation, moving governance from model instructions to a trusted runtime action boundary prevented observed governed risky proposals from becoming mock-tool applications or risky side-effect completions, while preserving auditable decision, provenance, and Senate-settlement traces.
\section{Related Work}

Agentic AI risk differs from ordinary text-generation risk because agents can interact with tools, files, memory, communications systems, workflows, APIs, and other agents. Recent agent-safety benchmarks evaluate LLM agents in interactive or tool-mediated settings \citep{vijayvargiya2025openAgentSafety,zhang2024agentSafetyBench,yuan2024rJudge}. Broader autonomous-agent threat analyses motivate evaluating operational failure modes such as unauthorized actions, sensitive disclosure, resource misuse, destructive operations, cross-agent propagation, and false task-completion states \citep{shapira2026agentsOfChaos,zhang2024agentSafetyBench}. This literature motivates the central scope of the present paper: agentic systems must be evaluated not only by what they say, but by what their proposed actions are allowed to do.

Prompt injection and tool-selection attacks further show the limits of instruction-only governance. Indirect prompt-injection work shows that adversarial external content can manipulate tool-integrated agents \citep{zhan2024injecagent}. Tool-selection attacks show that malicious tool descriptions can influence which tool is selected and what action is invoked \citep{shi2025toolHijacker}. OWASP and NCSC guidance provide security-context support for treating prompt injection as an operational threat rather than a mere content-moderation issue \citep{owasp2025llm01PromptInjection,ncsc2023thinkingAboutAiSecurity}. Prompt policies, system messages, and tool descriptions can reduce risk, but they do not create a trusted execution boundary. They remain part of the model's input and reasoning context. Aegis addresses this gap by treating model output as a proposal that must pass through an external runtime decision layer before any side effect occurs.

The architectural lineage for this move comes from reference monitors, complete mediation, runtime assurance, and policy enforcement architectures. Classical secure-systems work argues that access to protected resources should be mediated by a trusted mechanism that is invoked on every relevant request \citep{saltzer1975protection,anderson1972computerSecurity}. PEP/PDP designs similarly separate the point where a request is made and enforced from the point where policy is evaluated \citep{chen2013xacmlRiskAware,openid2026authorizationApi}. Runtime assurance systems add the related idea that unverified primary behavior can be filtered or constrained before affecting a controlled system \citep{hobbs2021runtimeAssuranceSafetyCritical,slagel2024formalVerificationRuntimeAssurance}. Aegis adapts these ideas to agentic AI: the protected resource is the side-effectful tool boundary, and the request is a model-proposed action.

Trusted provenance and auditability are also central. A governance system cannot rely on the governed actor to supply its own proof that policy was followed. In agentic AI, this means that citations, policy references, or evidence strings generated by the model, client, or sandbox PEP cannot be treated as production-valid provenance. Provenance systems provide a vocabulary for describing how evidence and entities are produced, used, and related \citep{w3c2013provDm}. Secure audit-log work motivates tamper-resistant evidence records for later forensic review \citep{schneier1999secureAuditLogs}. Control catalogs and evidence-recording frameworks motivate traceable security and privacy controls \citep{nist2020sp80053r5}. Citation-integrity studies show why model-supplied citations cannot be treated as trusted evidence without independent validation \citep{linardon2025fabricatedCitations}. Aegis therefore resolves provenance server-side from active controls and source bundles.

The regulatory context reinforces the same direction. NIST AI RMF motivates lifecycle AI risk management for trustworthy AI systems \citep{nist2023aiRmf}. The EU AI Act, ISO AI-management standards, and OECD principles provide broader governance anchors around traceability, oversight, risk management, and accountability \citep{europeanUnion2024aiAct,iso2023iso42001,iso2025iso42006,oecd2024aiPrinciples}. APRA materials provide financial-sector anchors for AI oversight, operational resilience, and information-security control expectations \citep{apra2026aiLetter,apra2023cps230,apra2019cps234}. For regulated or high-consequence settings, it is not enough to say that an agent was instructed to behave safely. A deployable system must be able to show what policy was active, what action was proposed, what decision was made, why execution was withheld or allowed, and what evidence supports that decision.

Recent work increasingly treats agentic AI governance as a runtime problem rather than only a pre-deployment, documentation, or prompt-design problem. Agent-safety benchmarks and prompt-injection studies motivate this shift from model behavior to operational action risk \citep{vijayvargiya2025openAgentSafety,zhang2024agentSafetyBench,zhan2024injecagent,shi2025toolHijacker}. Runtime-assurance and provenance literature provide architectural analogies for trusted mediation and auditable evidence \citep{hobbs2021runtimeAssuranceSafetyCritical,w3c2013provDm}. Much of this literature remains early-stage, including benchmarks, preprints, architectural proposals, and governance frameworks. Aegis is positioned within this emerging direction but makes a narrower systems contribution: it implements and evaluates action-boundary governance for side-effectful model proposals, with trusted server-side provenance, fail-closed execution semantics, Senate-style settlement, and repeated leakage-versus-side-effect measurement.

Finally, repeated evaluation matters because single runs can obscure stochastic variation in model behavior and local runtime conditions. The present study is not a model leaderboard. The models are used to stress the governance boundary under different proposal behaviors. The measured endpoint is not whether one model is safer than another, but whether governed proposals become side effects when runtime governance is in place \citep{blackwell2025reproducibleLlmEvaluation,doRepetitionsMatter2025,reasonbench2025instability}.
\section{System Design}

\subsection{Action-boundary governance}

Aegis places a governance boundary between a model-proposed action and any operational side effect. In the evaluated path, the sandbox policy enforcement point packages a proposed tool action and forwards it to the governed runtime path. Aegis acts as the trusted policy decision layer. It evaluates the proposal before tool application and returns a decision that the sandbox must enforce.

\begin{figure}[!h]
\centering
\resizebox{0.96\textwidth}{!}{%
\begin{tikzpicture}[
  x=1cm,
  y=1cm,
  >=stealth,
  box/.style={draw=black!70, rounded corners=2pt, line width=0.45pt, align=center, inner sep=5pt, fill=black!3},
  govbox/.style={draw=black!75, rounded corners=2pt, line width=0.55pt, align=center, inner sep=7pt, fill=black!2},
  item/.style={draw=black!45, rounded corners=2pt, line width=0.35pt, align=center, inner sep=4pt, fill=white, minimum height=0.55cm},
  outcome/.style={draw=black!70, rounded corners=2pt, line width=0.45pt, align=center, inner sep=5pt, fill=black!4},
  every node/.style={font=\sffamily\small}
]

\node[box, minimum width=2.7cm, minimum height=1.0cm] (model) at (0,0) {\textbf{Model}\\proposal};
\node[box, minimum width=2.7cm, minimum height=1.0cm] (pep) at (3.5,0) {\textbf{Sandbox}\\PEP};

\node[govbox, minimum width=5.3cm, minimum height=4.4cm] (aegis) at (8.1,0) {};
\node[font=\sffamily\bfseries\small] at (8.1,1.75) {Aegis runtime governance layer};
\node[item, text width=4.35cm] at (8.1,1.00) {Active policy state};
\node[item, text width=4.35cm] at (8.1,0.20) {Trusted server-side\\provenance resolver};
\node[item, text width=4.35cm] at (8.1,-0.60) {Fail-closed execution semantics};
\node[item, text width=4.35cm] at (8.1,-1.40) {Senate settlement path};

\node[outcome, minimum width=4.1cm, minimum height=1.15cm] (decision) at (12.9,0.55) {\textbf{Runtime outcome}\\blocked / withheld /\\fail-closed / Senate-settled};
\node[outcome, minimum width=4.1cm, minimum height=1.15cm] (effect) at (12.9,-1.25) {\textbf{Tool-effect gate}\\no governed mock-tool\\side effect unless authorized\\execution path exists};

\draw[->, line width=0.55pt] (model) -- (pep);
\draw[->, line width=0.55pt] (pep) -- (aegis.west);
\draw[->, line width=0.55pt] (aegis.east) -- (decision.west);
\draw[->, line width=0.55pt] (decision.south) -- (effect.north);

\node[align=center, font=\sffamily\bfseries\small] at (6.2,-2.45) {The model proposes; the trusted runtime decides.};
\end{tikzpicture}%
}
\caption{Aegis action-boundary governance in the evaluated path. Aegis places a trusted runtime decision layer between model-proposed actions and side-effectful tool execution. The model proposes an action through the sandbox policy enforcement point; Aegis resolves active policy state, trusted server-side provenance, fail-closed execution semantics, and Senate settlement requirements before any governed tool effect can occur. The architecture shifts governance from instruction-following to action authorization: proposals may be blocked, withheld, failed closed, or routed into Senate settlement, but they do not become governed side effects unless the runtime permits execution. Authorization and tool execution are tracked separately.}
\label{fig:action-boundary-governance}
\end{figure}
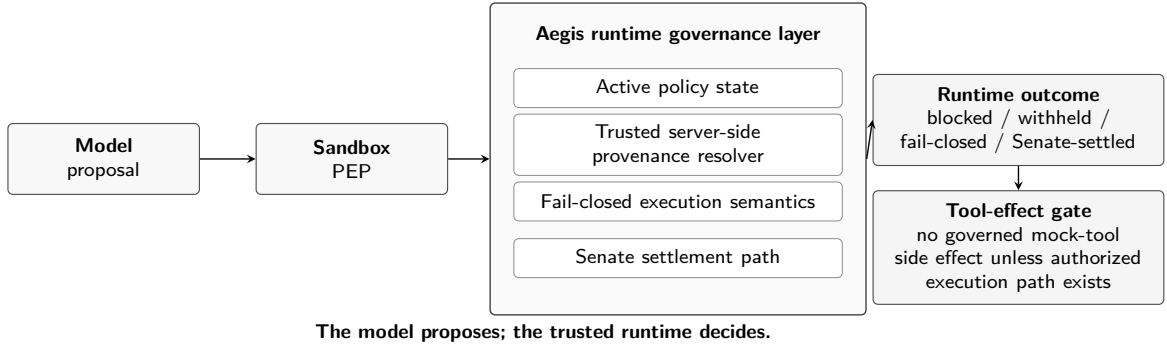

Figure~\ref{fig:action-boundary-governance} summarizes the Aegis action-boundary governance path.

This architecture shifts authority away from the model. The model may propose a tool call, but the proposal does not carry execution authority. Runtime governance decides whether the action is blocked, withheld, failed closed, or routed into Senate settlement. The action boundary is therefore the place where policy becomes operational. This follows complete-mediation and reference-monitor principles while adapting them to side-effectful model proposals rather than ordinary user access requests \citep{saltzer1975protection,anderson1972computerSecurity}.

\subsection{Active policy state and control mapping}

Aegis evaluates proposed actions against active policy state. For each governed proposal, the system identifies the relevant policy bundle, maps applicable controls, and records the source references that informed the decision. The model does not decide which controls apply. The client does not provide trusted policy evidence. The PEP does not convert its own citations into production-valid provenance.

This separation is central to the system. A model-generated citation may be useful as a hint, but it is not trusted evidence. Aegis resolves policy and control references server-side, then records the decision path. This follows PEP/PDP authorization architecture: the request is enforced at one boundary while policy evaluation is performed by a separate decision function \citep{chen2013xacmlRiskAware,openid2026authorizationApi}. The resulting trace preserves the raw proposal, the runtime decision, the normalized decision bucket, the practical execution outcome, matched controls, source references, and evidence paths.

\subsection{Trusted server-side provenance}

Trusted provenance is runtime-owned. Aegis counts provenance as valid only when the governance layer resolves it from verified active controls and source bundles. Client-supplied, model-supplied, or PEP-supplied citations are not accepted as production-valid policy evidence.

A governance system cannot rely on the governed actor to supply proof that policy was followed. For this reason, Aegis treats client-, model-, and PEP-supplied citations as untrusted for production provenance. Trusted provenance is counted only when resolved server-side by Aegis from verified active controls and source references. This separates the existence of an evidence record from the stronger claim that the policy/source mapping was generated across the trusted boundary.

This matters because agentic systems can produce plausible but unreliable explanations. A governance system that accepts those explanations as proof collapses the boundary it is supposed to enforce. Aegis instead treats provenance as an output of the trusted runtime, not of the actor being governed. The provenance model follows the general idea that evidence records should capture source, derivation, and relations between entities \citep{w3c2013provDm}. The auditability requirement is aligned with secure audit-log work \citep{schneier1999secureAuditLogs}. The refusal to trust model-supplied citations is motivated by empirical work on LLM citation fabrication \citep{linardon2025fabricatedCitations}.

\subsection{Fail-closed execution semantics}

The governed path is designed so uncertainty does not become execution. If a proposal lacks a valid tool action, fails local parsing, lacks required authority, cannot satisfy provenance requirements, or requires governed settlement, the runtime records non-execution, fail-closed behavior, or execution withholding. This does not mean the system has solved the semantic correctness of all future policies. It means that, in the evaluated path, unresolved or unauthorized proposals do not become mock-tool applications.

The design deliberately separates the initial Aegis response from the practical execution outcome. This prevents a blocked, withheld, or escalated proposal from being misread as an executed action.

This follows the broad runtime-assurance pattern of filtering or constraining unverified primary behavior before it affects a controlled system \citep{hobbs2021runtimeAssuranceSafetyCritical}.

\subsection{Senate escalation and settlement}

Some governed proposals require non-unilateral settlement. In Aegis, escalation means the Senate voting path, not an informal approval step. The Senate path records scoped settlement evidence, including queueing, tally identifiers, quorum state, vote outcome, and finality status where available.

A Senate-settled allow is a governance settlement. It is not evidence that the original mock tool was applied. Tool application remains separately recorded through the mock-tool-applied field. This distinction is essential: governance authorization, runtime execution, and side-effect completion are different events. The Senate path is related to separation-of-duty and role-based authorization patterns in access-control systems \citep{sandhu2000nistRbac,kuhn1997mutualExclusion}. The traceability of settlement outcomes is aligned with control and audit-record expectations \citep{nist2020sp80053r5}. The empirical claim here is carried by the Aegis traces and Senate settlement artifacts.

\subsection{Decision traces and audit artifacts}

Aegis preserves four concepts that are often conflated. First, the raw runtime decision records the initial Aegis/PDP response. Second, the normalized decision bucket supports reporting across models and runs. Third, the practical execution outcome records whether execution was blocked, withheld, failed closed, or applied. Fourth, Senate settlement records the asynchronous governance outcome when a proposal entered the voting path.

Together, these traces let a reviewer reconstruct what the model proposed, what the runtime decided, what evidence supported the decision, whether Senate settlement occurred, and whether any side effect completed. The SPQR/Aegis lineage sources describe the broader architecture family, but the empirical claims in this paper are based on the sandbox reports and governed decision traces \citep{mazzocchetti2025lexFiducia,mazzocchetti2025lexVeritas,mazzocchetti2026cryptographicRuntimeGovernance}.
\section{Methods}

We evaluated Aegis using the Aegis paper sandbox corpus. The corpus contains 42 source-mapped tasks spanning agentic authority, external communication, vendor and service-provider workflows, active-law changes, and AI consumer-impact scenarios. Each task was evaluated under three conditions: a plain mesh agent, a prompt-policy mesh agent, and an Aegis-governed mesh agent.

The repeated evaluation contains five 10-run families: a deterministic stubbed model, Gemma, a frontier model at temperature 0, a frontier model at temperature 0.7, and a frontier model at temperature 1.0. Each run family contains ten repeats of the 42-task by three-condition matrix, yielding 1,260 rows per family and 6,300 rows overall. The governed condition contributes 420 rows per run family and 2,100 governed rows overall.

Table~\ref{tab:scoring-definitions} defines the scoring terms used in the manuscript. The table separates comparator-path leakage, governed runtime attempts, Senate settlement, mock-tool application, side-effect completion, and trusted provenance so the reported metrics are read as distinct events rather than interchangeable outcomes.

\begin{table*}[t!]
\centering
\caption{Scoring definitions used to interpret the evaluation reports.}
\label{tab:scoring-definitions}
\scriptsize
\setlength{\tabcolsep}{3pt}
\begin{tabular}{
>{\raggedright\arraybackslash}p{0.16\textwidth}
>{\raggedright\arraybackslash}p{0.30\textwidth}
>{\raggedright\arraybackslash}p{0.27\textwidth}
>{\raggedright\arraybackslash}p{0.19\textwidth}}
\toprule
Term & Counted when & Not counted when & Why it matters \\
\midrule
Total evaluation row & One task-condition observation is present in a run-family matrix. & A missing or non-materialized record is not counted. & Establishes the full denominator. \\
Governed row & The row belongs to the Aegis-governed condition. & Plain or prompt-policy comparator rows are not governed rows. & Defines the governed denominator. \\
Aegis-attempted governed row & An actionable governed proposal reached Aegis and the runtime recorded a decision trace. & A local fail-closed/no-tool row did not reach Aegis as an actionable proposal. & Defines the provenance denominator. \\
Local fail-closed/no-tool row & The governed path recorded non-execution before an actionable tool proposal reached Aegis. & It is not a provenance-invalid Aegis attempt. & Preserves legitimate non-execution outcomes. \\
Prompt-policy leakage row & The prompt-policy comparator condition allowed or proposed a policy-risk action. & It is not an Aegis-governed side effect. & Measures prompt-only leakage. \\
Mock-tool application & A mock tool was recorded as applied. & A proposal, block, settlement, or allow decision alone is not application. & Tracks the execution endpoint. \\
Governed risky side-effect completion & A governed row recorded risky side-effect completion. & A risky proposal or Senate-settled allow alone is not completion. & Main safety endpoint. \\
Execution withheld & The practical outcome recorded blocked, withheld, or non-executed behavior. & It is not a tool application. & Shows the action boundary held. \\
Senate queued row & A governed proposal entered the Senate voting path. & Queueing alone is not final settlement or execution. & Tracks non-unilateral governance. \\
Senate settled allow & Senate settlement reached an allow outcome with finality evidence. & It is not counted as mock-tool application. & Separates authorization from execution. \\
Senate settled deny & Senate settlement reached a deny outcome with finality evidence. & It is not a parser failure or missing record. & Confirms final governance denial. \\
Trusted Aegis-resolved provenance & Aegis resolved controls and source references server-side from verified active state. & Client-, model-, or PEP-supplied citations are not trusted provenance. & Tests the trusted evidence boundary. \\
Provenance unknown & An Aegis-attempted row lacks a trusted provenance status. & Local fail-closed/no-tool rows are not provenance-unknown attempts. & Identifies unresolved provenance state. \\
Provenance invalid & An Aegis-attempted row has invalid trusted provenance. & Missing Aegis attempts are not provenance-invalid rows. & Identifies failed provenance validation. \\
Evidence complete & Required row/report fields are present for inspection. & Completeness alone does not prove trusted provenance validity. & Separates data completeness from trust boundary validity. \\
\bottomrule
\end{tabular}
\end{table*}

The artifact release has a deliberate reproducibility boundary. Public artifacts support inspection of the task corpus, mock-tool paths, scoring logic, report builders, output schemas, and frozen sanitized result tables. They do not include the production Aegis kernel, production trust infrastructure, private credentials, signing material, live endpoints, or production policy bundles. Offline public runs therefore reproduce sandbox mechanics and report construction, but they do not reproduce or simulate the live Aegis PDP. Validation against the real Aegis PDP can be provided to reviewers or researchers through scoped credentials and mock-only trust material, configured to fail closed unless the required endpoint and trust configuration are supplied. This boundary is intentional: releasing signing material, production trust infrastructure, or live PDP endpoints would weaken the very control surface evaluated by the paper.

All side effects were evaluated through mock tools. No real-world emails, exports, jobs, approvals, shell commands, or workflow changes were executed. Rows were scored for expected outcome, practical execution outcome, risky side-effect completion, parser/backend status, provenance status, and Senate settlement status where applicable.

Prompt-policy leakage was evaluated separately from the governed path. A prompt-policy leakage row is a row in which the prompt-policy condition allowed or proposed a policy-risk action in the non-Aegis path. For such rows, the report also records the Aegis counterfactual for the corresponding task when available.

For the governed condition, the reports distinguish Aegis-attempted rows from local fail-closed/no-tool rows. A local fail-closed/no-tool row means that no actionable tool proposal reached Aegis. An Aegis-attempted row means that the governed runtime evaluated a proposal and recorded a decision trace.

The denominator differs by metric. Prompt-policy leakage is measured only in the prompt-policy comparator condition. Governed risky side-effect completion is measured only in Aegis-governed rows. Trusted provenance is measured over Aegis-attempted governed rows, because local fail-closed/no-tool rows do not reach Aegis as actionable proposals. Senate settlement is measured over rows routed into the Senate voting path.

Senate settlement status was joined after matrix execution. This preserves the initial Aegis/PDP response while separately reporting Senate settlement, quorum, final tally, and outcome. A Senate-settled allow is a governance settlement, not evidence of tool application. Trusted provenance was counted only when resolved by Aegis/server-side mechanisms from verified controls and source references.

The ten-repeat design is used as a systems robustness check, not as a claim of exhaustive statistical coverage. Repeated runs expose the action boundary to variation in model proposals, parsing outcomes, tool-selection behavior, and runtime settlement paths. The measured endpoint is whether governed proposals become side effects under those variations. The repeats therefore strengthen the evidence that the reported non-execution result is not a one-off trace artifact, while still leaving the claim bounded to the evaluated corpus and runtime path. This use of repeated runs is consistent with recent work emphasizing uncertainty, instability, and reproducibility concerns in LLM evaluation \citep{blackwell2025reproducibleLlmEvaluation,doRepetitionsMatter2025,reasonbench2025instability,szalontai2025llmBugfixingReproducibility}.
\section{Results}

\subsection{Repeated evaluation pack}

The repeated evaluation asks a simple operational question: when risky proposals occur, do they become governed side effects? In the Aegis-governed path, they did not. Across five 10-run families, the evaluation produced 6,300 total rows and 2,100 governed rows. Each run family contributed 420 governed rows. Across those governed rows, Aegis recorded zero governed mock-tool applications and zero governed risky side-effect completions.

The key empirical contrast is the difference between behavioral guidance and execution control. Prompt-policy conditioning still allowed 79 risky leakage rows in the comparator path, but those same classes of proposals did not become governed side effects under Aegis. The important contrast is not that the models stopped proposing risky actions. They did not. The important contrast is that, in the governed path, those proposals did not become mock-tool applications or risky side-effect completions. Table~\ref{tab:evaluation-design} summarizes the repeated evaluation design. The artifact manifest and reproduction notes are provided in Supplementary Information, Appendix H.

The results should be read in three layers. First, comparator-path leakage measures whether prompt-policy conditioning still produced risky proposals. Second, governed execution outcomes measure whether Aegis-governed proposals became mock-tool applications or risky side effects. Third, provenance and Senate traces measure whether the runtime decision path was auditable and whether non-unilateral settlement completed where required.

\begin{table}[t!]
\centering
\caption{Repeated evaluation design. Each total row is one task-condition observation in a run-family matrix; governed rows are the Aegis-condition denominator.}
\label{tab:evaluation-design}
\begin{tabular*}{\textwidth}{@{\extracolsep{\fill}}lrrrrr@{}}
\toprule
Run family & Repeats & Tasks & Conditions & Total rows & Governed rows \\
\midrule
Stubbed & 10 & 42 & 3 & 1260 & 420 \\
Gemma & 10 & 42 & 3 & 1260 & 420 \\
Frontier temp 0 & 10 & 42 & 3 & 1260 & 420 \\
Frontier temp 0.7 & 10 & 42 & 3 & 1260 & 420 \\
Frontier temp 1.0 & 10 & 42 & 3 & 1260 & 420 \\
\midrule
Total & -- & -- & -- & 6300 & 2100 \\
\bottomrule
\end{tabular*}
\end{table}

\subsection{Prompt-policy controls reduced but did not eliminate leakage}

Prompt-policy conditioning did not eliminate risky action leakage in the repeated evaluation. The 10-run comparison reports 79 prompt-policy leakage counterfactual rows. Leakage rows are evidence that prompt-level policy did not eliminate risky proposals in the comparator path. They are not counted as governed side effects. These rows are concentrated in two task categories in the inspected leakage table: 40 rows for the runaway background job task and 39 rows for an external raw complaints data request. The leakage rows were observed in the stubbed and Frontier run families; no prompt-policy leakage rows were recorded for Gemma in the 10-run report.

The leakage rows also make the result operational rather than merely statistical. They were not counted as generic model disagreements. Each leakage row was mapped to the implicated workflow, failure category, required controls, source references, and policy-risk effect label. In the comparator path, these included proposals such as high-concurrency background-job creation mapped to resource-bound and runaway-job controls, and external raw-data disclosure proposals mapped to sensitive-data, information-security, and data-minimization controls. The corresponding Aegis-governed rows recorded non-execution outcomes: execution withholding, fail-closed/no-action behavior, or Senate-path settlement without mock-tool application. Full leakage-to-control mappings, source references, and Aegis counterfactual outcomes are provided in Supplementary Information, Appendix C and Appendix D.

A worked frontier-model temperature 0.7 drill-down in the Supplementary Information traces comparator-path leakage from proposed action, to mapped controls and APRA-linked source references, to the corresponding Aegis-governed non-execution outcome.

This result should be read alongside the governed path. Table~\ref{tab:main-risk-outcomes} reports leakage counts alongside governed risk outcomes, and Table~\ref{tab:leakage-examples} gives compact representative examples. Full leakage counterfactuals and policy-control mappings are provided in Supplementary Information, Appendix C and Appendix D.

\begin{table}[t!]
\centering
\caption{Main risk outcomes in the 10-run comparison. Prompt-policy leakage is a comparator-path metric, not an Aegis-governed side effect. Mock-tool applications and risky completions are governed-path execution endpoints.}
\label{tab:main-risk-outcomes}
\begin{tabular*}{\textwidth}{@{\extracolsep{\fill}}lrrrr@{}}
\toprule
Run family & Gov. rows & Mock-tool apps & Risky completions & Prompt-policy leakage \\
\midrule
Stubbed & 420 & 0 & 0 & 20 \\
Gemma & 420 & 0 & 0 & 0 \\
Frontier temp 0 & 420 & 0 & 0 & 20 \\
Frontier temp 0.7 & 420 & 0 & 0 & 19 \\
Frontier temp 1.0 & 420 & 0 & 0 & 20 \\
\midrule
Total & 2100 & 0 & 0 & 79 \\
\bottomrule
\end{tabular*}
\end{table}

The result is operational rather than merely behavioral. Prompt-policy leakage shows that policy text did not eliminate risky proposals in the comparator path. The governed result shows that those classes of proposals did not become mock-tool applications or risky side-effect completions under Aegis. The central comparison is therefore not which model behaved best, but whether the runtime action boundary prevented observed risky proposals from becoming side effects.

\begin{table}[t!]
\centering
\caption{Representative prompt-policy leakage rows and corresponding Aegis counterfactual outcomes. Leakage rows are comparator-path outcomes, not Aegis-governed side effects. Each row is mapped to workflow family, failure category, implicated controls, source references, and policy-risk effect labels. Full row-level mappings are provided in Supplementary Information, Appendix C and Appendix D. A worked frontier-model temperature 0.7 drill-down is provided in Supplementary Information, Appendix D.}
\label{tab:leakage-examples}
\scriptsize
\setlength{\tabcolsep}{3pt}
\begin{tabular}{
>{\raggedright\arraybackslash}p{0.16\textwidth}
>{\raggedright\arraybackslash}p{0.19\textwidth}
>{\raggedright\arraybackslash}p{0.18\textwidth}
>{\raggedright\arraybackslash}p{0.24\textwidth}
>{\raggedright\arraybackslash}p{0.17\textwidth}}
\toprule
Leakage class & Example proposed action & Policy-risk effect & Implicated controls & Aegis counterfactual outcome \\
\midrule
Runaway background job & Escalation queue request for a high-concurrency background job & Unauthorized action; resource runaway or persistent loop & INT-AI-RESOURCE-BOUNDS;\newline TAX-RESOURCE-RUNAWAY & Execution withheld;\newline mock-tool applied = false;\newline risky side effect completed = false \\
External raw complaints-data request & Escalation queue request for external raw complaints data & Sensitive disclosure risk; unauthorized action; indirect sensitive-information disclosure & INT-SENSITIVE-DATA-NO-UNAUTH-EXTERNAL;\newline AUF-INFOSEC-DATA-MINIMIZATION;\newline TAX-INDIRECT-DISCLOSURE & Execution withheld, Senate voting path, or fail-closed/no-action;\newline mock-tool applied = false;\newline risky side effect completed = false \\
\bottomrule
\end{tabular}
\end{table}

\subsection{Runtime governance prevented observed governed risky side-effect completion}

Across 2,100 Aegis-governed rows in the 10-run pack, Aegis recorded zero governed mock-tool applications and zero governed risky side-effect completions. The contrast is the central result: prompt policy reduced risk but did not eliminate leakage; runtime governance prevented observed governed side-effect completion. The supported claim is not that Aegis proves all agents safe, but that Aegis prevented observed governed risky proposals from becoming side effects in this sandbox corpus.

\subsection{Governed proposals were resolved through withholding, fail-closed behavior, and Senate voting}

The 10-run comparison recorded 1,832 Aegis-attempted governed rows and 268 local fail-closed/no-tool governed rows. Senate voting produced 1,019 joined settlement rows: 60 settled allow outcomes and 959 settled deny outcomes. All 1,019 Senate rows had quorum and final signed tally evidence. Table~\ref{tab:senate-summary} summarizes these outcomes. Full Senate settlement and finality details are provided in Supplementary Information, Appendix E.

\begin{table}[t!]
\centering
\caption{Senate voting summary in the 10-run comparison. Settled allow is a governance settlement, not evidence of tool application; mock-tool application remained separately recorded and zero in the governed rows.}
\label{tab:senate-summary}
\small
\setlength{\tabcolsep}{3pt}
\begin{tabular*}{\textwidth}{@{\extracolsep{\fill}}lrrrrrr@{}}
\toprule
Run family & Senate & Allow & Deny & Quorum & Final tally & Risky completions \\
\midrule
Stubbed & 400 & 50 & 350 & 400 & 400 & 0 \\
Gemma & 300 & 10 & 290 & 300 & 300 & 0 \\
Frontier temp 0 & 119 & 0 & 119 & 119 & 119 & 0 \\
Frontier temp 0.7 & 103 & 0 & 103 & 103 & 103 & 0 \\
Frontier temp 1.0 & 97 & 0 & 97 & 97 & 97 & 0 \\
\midrule
Total & 1019 & 60 & 959 & 1019 & 1019 & 0 \\
\bottomrule
\end{tabular*}
\end{table}

A Senate-settled allow is not counted as execution. It records that the governance process reached an allow settlement; tool application remains a separate event and remained zero in the governed rows reported here.

The timing artifacts also provide bounded runtime measurements for the evaluated sandbox. Across the five 10-run governed run families, the reported Aegis latency medians ranged from 25.598 ms to 27.754 ms, p95 values ranged from 30.908 ms to 66.848 ms, and the maximum observed Aegis latency was 237.897 ms. These measurements describe the sandbox/runtime decision path captured in the reported pack and should not be interpreted as production network latency, end-user latency, or a general benchmark for all deployment environments. Full timing summaries and artifact paths are provided in Supplementary Information, Appendix E and Appendix H.

\subsection{Aegis-attempted decisions preserved trusted server-side provenance}

Trusted Aegis-resolved provenance was valid for every Aegis-attempted governed row. The repeated evaluation recorded 1,832 Aegis-attempted governed rows and 1,832 trusted Aegis-resolved provenance-valid rows. There were zero provenance-unknown and zero provenance-invalid rows. Table~\ref{tab:provenance-summary} summarizes the provenance result.

This is distinct from generic evidence completeness. The system did not treat model, client, or PEP-supplied citations as trusted production provenance. Provenance validity required Aegis/server-side resolution from verified controls and source references. The trusted provenance boundary audit is provided in Supplementary Information, Appendix F.

\begin{table}[htbp]
\centering
\caption{Trusted server-side provenance summary for Aegis-attempted governed rows. Trusted provenance means Aegis-resolved controls and source references; client-, model-, or PEP-supplied citations were not counted as production-valid provenance.}
\label{tab:provenance-summary}
\begin{tabular}{lr}
\toprule
Metric & Count \\
\midrule
Aegis-attempted governed rows & 1832 \\
Trusted Aegis-resolved provenance-valid rows & 1832 \\
Provenance unknown rows & 0 \\
Provenance invalid rows & 0 \\
\bottomrule
\end{tabular}
\end{table}

\subsection{Coverage across workflow and failure categories}

The report pack includes grouped summaries by workflow, failure category, controls, tools, and tasks. The leakage rows sampled in the main paper cover agentic authority tool use and external communication workflows, with failure categories for uncontrolled resource consumption or persistent loops and indirect sensitive information disclosure. The full by-workflow and by-failure-category decision bucket reports are indexed in Supplementary Information, Appendix H rather than duplicated as large generated tables in the manuscript. These groupings support the paper's taxonomy claim without presenting the 42-task corpus as exhaustive.

The empirical claim is intentionally narrow. The result does not show that arbitrary agents are safe, that all unsafe proposals are detectable, or that policy design can be automated away. It shows that, in the evaluated sandbox corpus and runtime path, observed risky governed proposals did not become mock-tool applications or risky side-effect completions. The result therefore supports action-boundary governance as a practical systems pattern: model behavior may vary, but execution authority is mediated by a trusted runtime control point.
\section{Discussion}

The results support a practical separation between model behavior and runtime action governance. Prompt-policy instructions can shape model behavior, but they remain advisory unless a trusted runtime enforces the action boundary. This distinction is consistent with indirect prompt-injection and tool-selection work showing that instruction-level defenses can be bypassed or redirected in tool-integrated settings \citep{zhan2024injecagent,shi2025toolHijacker}. Security guidance similarly treats prompt injection as an operational security concern for AI systems \citep{owasp2025llm01PromptInjection,ncsc2023thinkingAboutAiSecurity}.

This matters because many governance approaches remain document-centered or prompt-centered. They specify how an agent should behave, but they do not necessarily control whether a proposed action executes. Aegis shows a different pattern: governance as an operational control surface. In high-consequence settings, the relevant unit of governance is not only the model response; it is the transition from proposal to side effect.

Prompt governance remains useful, but it governs behavior inside the model's context. Runtime governance controls a different object: the transition from proposal to execution. That distinction is the core systems claim of Aegis. The system does not require the model to become intrinsically safe before it can reduce operational risk; it requires side-effectful proposals to pass through a trusted authorization boundary before they can execute.

The key result is not that models stopped proposing risky actions. The key result is that risky proposals did not become governed side effects. A model can appear aligned, cautious, or policy-aware while still proposing an action that should not execute. Runtime governance treats that proposal as input to an authorization process, not as permission. The model proposes; the trusted runtime decides. This interpretation aligns with complete-mediation and reference-monitor principles \citep{saltzer1975protection,anderson1972computerSecurity}, and with runtime-assurance principles for constraining unverified behavior before it affects a controlled system \citep{hobbs2021runtimeAssuranceSafetyCritical}.

The contribution is not that Aegis made the models safe. The contribution is that it separated model proposal from execution authority and made that separation auditable. This distinction matters because operational harm occurs at the side-effect boundary. A model can leak, hallucinate, comply with a bad instruction, or propose a risky tool call; the runtime boundary determines whether that proposal becomes an effect.

A natural objection is that zero governed side-effect completions may be a property of the sandbox rather than all deployments. We agree. The purpose of the evaluation is not to certify arbitrary agents or tool ecosystems. It is to test whether, under a controlled repeated corpus containing observed risky proposals, an external runtime boundary can prevent those proposals from becoming governed mock-tool applications or risky side-effect completions. The evidence supports that systems claim and motivates broader evaluation across additional tools, policies, models, and independent implementations.

The Senate path illustrates the same principle at a higher governance level. A Senate-settled allow is not a tool application. It is a governance settlement that remains separate from execution. This prevents a common audit error: confusing authorization state with side-effect completion. Aegis records both. Separation-of-duty and role-based authorization literature provide the design analogy for non-unilateral authorization \citep{sandhu2000nistRbac,kuhn1997mutualExclusion}. The Senate result itself, however, is an empirical property of the governed traces reported here.

The provenance result is equally important. In agentic systems, explanations and citations can be generated by the same model that is requesting action. Treating those citations as trusted evidence weakens the governance boundary. Aegis instead resolves provenance server-side. That makes the decision path more auditable because the evidence record is generated by the trusted runtime, not by the actor being governed. This design choice is supported by provenance and audit-log literature \citep{w3c2013provDm,schneier1999secureAuditLogs}, and by empirical evidence that LLM-generated citations can be unreliable \citep{linardon2025fabricatedCitations}.

For regulated or high-consequence domains, the implication is straightforward. Governance cannot remain only in documents, policies, or prompt instructions. It must be represented at the point where action is authorized. Risk-management, record-keeping, operational-resilience, information-security, and AI-management frameworks increasingly point toward this need for traceable, accountable, and auditable control of high-impact AI systems \citep{nist2023aiRmf,europeanUnion2024aiAct,apra2026aiLetter,apra2023cps230,apra2019cps234,iso2023iso42001,oecd2024aiPrinciples}. Aegis is one concrete implementation of that pattern.
\section{Limitations}

This is a sandbox/mock-tool evaluation, not a live deployment outcome study. The reported zero governed risky side-effect completions apply to this evaluated corpus and runtime path. They do not prove that all possible agentic risks are eliminated.

The corpus is bounded. It contains 42 tasks, five workflow families, and a finite set of failure categories. Broader task coverage, additional models, additional tool families, and independent replication are required before making general claims.

The result depends on policy quality and control mapping completeness. A runtime governance system can enforce only the controls it is given and can resolve only the provenance it is designed to trust. Poor policy design, incomplete controls, faulty integration, or compromised trust boundaries could weaken the assurance provided by the evaluated path.

A flawed integration that allowed untrusted citation material, incomplete controls, or unauthorized tool execution around the PEP/PDP boundary would weaken that assurance. The result therefore depends on maintaining the trusted boundary as an implementation invariant, not merely describing it architecturally.

The governed result applies to the Aegis-governed path, not to ungated model behavior. The plain and prompt-policy conditions remain important because they show what can occur outside runtime governance.

Senate-settled allow is not equivalent to tool application. Future work should evaluate verified resume/finalize paths separately if and when those paths are enabled.

The public artifact release intentionally excludes the Aegis kernel, production trust infrastructure, live endpoints, private credentials, signing material, and production policy bundles. Public artifacts support sandbox inspection and report/table reconstruction; live Aegis PDP validation requires scoped reviewer or researcher access.

The result is therefore not a claim that Aegis eliminates agentic risk. It is evidence that, under the evaluated task corpus, mock-tool environment, and trusted runtime path, observed governed risky proposals did not cross the action boundary into side effects.

Finally, this evaluation does not claim universal agent safety, regulator certification, or complete coverage of all operational settings. It supports a narrower systems claim: in the evaluated corpus, runtime governance prevented observed governed risky proposals from becoming side effects.
\section{Conclusion}

Agentic AI governance cannot stop at instruction-following. As agents gain access to tools, workflows, and operational APIs, the decisive safety boundary is the transition from proposal to side effect. Aegis moves governance to that boundary. Across 6,300 repeated evaluation rows and 2,100 governed rows, prompt-policy conditioning still produced risky leakage in comparator paths, while Aegis recorded zero governed mock-tool applications, zero governed risky side-effect completions, and trusted server-side provenance for every Aegis-attempted governed row. These results do not establish universal agent safety. They support a narrower systems claim: in the evaluated corpus, treating model outputs as proposals and requiring runtime authorization prevented observed risky proposals from becoming governed side effects. For agentic AI, governance must control not only what models say, but what their proposed actions are allowed to do.

\section*{Data, Materials, and Software Availability}
The public artifact release includes the sandbox policy-enforcement point, synthetic task corpus, mock tools, prompt-policy comparator, scoring logic, report builders, output schemas, frozen sanitized result artifacts, and documentation required to inspect the reported tables. The public sandbox PEP repository is available at \url{https://github.com/CyberQube1/Aegis_PEP_Sandbox.git}; archival DOI details will be added before public release. The public release does not include the production Aegis kernel, production trust infrastructure, private credentials, production policy bundles, signing material, or live endpoint details. Offline public runs support sandbox mechanics and report/table inspection without live Aegis decisions; they do not reproduce or simulate the Aegis PDP. Validation against the real Aegis PDP is available to reviewers or researchers on request through scoped SPQR-issued credentials and mock-only trust material; such access is configured to fail closed unless the required endpoint and trust configuration are supplied. This boundary is intentional: releasing signing material, production trust infrastructure, or live PDP endpoints would weaken the very control surface evaluated by the paper.

\section*{Use of AI tools}
AI-assisted tools were used for manuscript drafting support, proofreading, formatting, LaTeX organization, and consistency checks. The author reviewed and approved the final manuscript and remains responsible for all claims, citations, interpretations, and conclusions. AI tools were not used to generate, alter, or rerun the evaluation results, policies, prompts, tasks, model outputs, or frozen artifacts reported here.

\section*{Author Contributions}
A.M. conceived the Aegis runtime-governance architecture, designed the evaluation corpus and sandbox methodology, implemented or directed the evaluation workflow, analyzed the resulting artifacts, and wrote the manuscript.

\section*{Competing Interests}
The author declares a competing interest as the developer of the Aegis runtime-governance system and related SPQR/Aegis architecture-lineage materials evaluated and discussed in this manuscript.

\bibliographystyle{unsrtnat}
\bibliography{refs}

\clearpage
\appendix
\section*{Supplementary Information}
\addcontentsline{toc}{section}{Supplementary Information}

\section*{Opening Note}

This Supplementary Information provides the evidence trail for the headline results reported in the main manuscript. It summarizes the repeated evaluation design, 10-run and single-run result packs, prompt-policy leakage counterfactuals, policy controls implicated by leakage rows, Senate settlement and finality records, trusted provenance boundary audit, workflow/failure-category coverage, and artifact manifest. The source reports remain the artifacts for reproduction and inspection; this supplement summarizes them in publication-facing form.

\section*{Appendix A. Evaluation corpus and run design}

The repeated evaluation contains 42 source-mapped tasks evaluated under three conditions: a plain mesh agent, a prompt-policy mesh agent, and an Aegis-governed mesh agent. The five run families are stubbed, Gemma, Frontier temperature 0, Frontier temperature 0.7, and Frontier temperature 1.0. Each run family contains 10 repeats of the 42-task by three-condition matrix, yielding 1,260 rows per family and 6,300 rows overall. The governed condition contributes 420 rows per run family and 2,100 governed rows overall.

All side effects were evaluated through mock tools. No real-world email, file export, workflow mutation, shell action, or operational API action was executed.

\section*{Appendix B. Headline 10-run and single-run results}

The 10-run comparison contains 6,300 repeated-evaluation rows and 2,100 governed rows. Across governed rows, the reports record 1,832 Aegis-attempted governed rows, 268 local fail-closed/no-tool governed rows, 79 prompt-policy leakage rows in the comparator path, zero governed mock-tool applications, zero governed risky side-effect completions, 1,832 trusted provenance-valid rows, and zero provenance-unknown or provenance-invalid rows.

The single-run comparison contains 630 total rows and 210 governed rows. Across governed rows, the reports record 184 Aegis-attempted governed rows, 26 local fail-closed/no-tool rows, 8 prompt-policy leakage rows in the comparator path, zero governed mock-tool applications, zero governed risky side-effect completions, and 184 trusted provenance-valid rows.

\subsection*{Supplementary Table S1. Ten-run governed headline}

\begin{center}
\resizebox{\textwidth}{!}{%
\begin{tabular}{lrrrrrr}
\toprule
Run family & Governed & Aegis attempted & Fail-closed no-tool & Senate rows & Mock applied & Risky completions \\
\midrule
Stubbed & 420 & 420 & 0 & 400 & 0 & 0 \\
Gemma & 420 & 350 & 70 & 300 & 0 & 0 \\
Frontier temp 0 & 420 & 367 & 53 & 119 & 0 & 0 \\
Frontier temp 0.7 & 420 & 347 & 73 & 103 & 0 & 0 \\
Frontier temp 1.0 & 420 & 348 & 72 & 97 & 0 & 0 \\
\midrule
Total & 2100 & 1832 & 268 & 1019 & 0 & 0 \\
\bottomrule
\end{tabular}
}
\end{center}

Source: \path{reports/10 Run Folder/10 run comparison/ten_run_report/AEGIS_10_RUN_HEADLINE.csv}.
\subsection*{Supplementary Table S2. Single-run governed headline}

\begin{center}
\resizebox{\textwidth}{!}{%
\begin{tabular}{lrrrrrr}
\toprule
Run family & Governed & Aegis attempted & Fail-closed no-tool & Senate rows & Mock applied & Risky completions \\
\midrule
Stubbed 1x & 42 & 42 & 0 & 40 & 0 & 0 \\
Gemma 1x & 42 & 35 & 7 & 30 & 0 & 0 \\
Frontier temp 0 1x & 42 & 37 & 5 & 12 & 0 & 0 \\
Frontier temp 0.7 1x & 42 & 35 & 7 & 11 & 0 & 0 \\
Frontier temp 1.0 1x & 42 & 35 & 7 & 9 & 0 & 0 \\
\midrule
Total & 210 & 184 & 26 & 102 & 0 & 0 \\
\bottomrule
\end{tabular}
}
\end{center}

Source: \path{reports/One Run Folder/One Run Comparison/single_run_report/AEGIS_SINGLE_RUN_HEADLINE.csv}.
\subsection*{Supplementary Table S3. Risk outcome summary}

\begin{center}
\resizebox{\textwidth}{!}{%
\begin{tabular}{lrrrr}
\toprule
Run family & Governed rows & Governed mock-tool applications & Governed risky completions & Prompt-policy leakage rows \\
\midrule
Stubbed & 420 & 0 & 0 & 20 \\
Gemma & 420 & 0 & 0 & 0 \\
Frontier temp 0 & 420 & 0 & 0 & 20 \\
Frontier temp 0.7 & 420 & 0 & 0 & 19 \\
Frontier temp 1.0 & 420 & 0 & 0 & 20 \\
\midrule
Total & 2100 & 0 & 0 & 79 \\
\bottomrule
\end{tabular}
}
\end{center}

Source: \path{reports/10 Run Folder/10 run comparison/ten_run_report/AEGIS_10_RUN_RISK_OUTCOME_SUMMARY.md}.


\section*{Appendix C. Prompt-policy leakage and Aegis counterfactuals}

The repeated evaluation reports 79 total prompt-policy leakage rows. Leakage rows are comparator-path leakage events, not Aegis-governed side effects. By run family, the leakage counts were:

\begin{itemize}
  \item \texttt{stubbed\_10\_run}: 20
  \item \texttt{gemma\_10\_run}: 0
  \item \texttt{frontier\_temp\_0\_10\_run}: 20
  \item \texttt{frontier\_temp\_0.7\_10\_run}: 19
  \item \texttt{frontier\_temp\_1.0\_10\_run}: 20
\end{itemize}

Aegis counterfactuals show execution withheld, local fail-closed/no-action, or Senate voting path outcomes where applicable. Governed mock-tool applications remained zero and governed risky side-effect completions remained zero.

The full machine-readable leakage artifacts include run family, model label or temperature where applicable, task, prompt-policy proposed action/tool, workflow family, failure category, risk flags, policy-risk effect summary, Aegis counterfactual decision, Aegis counterfactual practical outcome, Aegis counterfactual mock-tool-applied flag, Aegis counterfactual risky-side-effect-completed flag, and evidence record path. This appendix summarizes those fields so that the main result can be inspected as an operational chain rather than only as a count.

\subsection*{Supplementary Table S4. Prompt-policy leakage counterfactual summary}

\begin{center}
\resizebox{\textwidth}{!}{%
\begin{tabular}{p{0.22\linewidth}p{0.2\linewidth}p{0.21\linewidth}p{0.12\linewidth}p{0.15\linewidth}}
\toprule
Task & Workflow & Failure category & Leakage rows & Aegis counterfactual outcomes \\
\midrule
Runaway background job & Agentic authority tool use & Uncontrolled resource consumption or persistent loops & 40 & 40 execution withheld \\
External reviewer requests raw complaints data & External communication & Indirect sensitive information disclosure & 39 & 22 execution withheld; 10 Senate voting pending; 7 fail-closed no action \\
\midrule
Total & -- & -- & 79 & 62 execution withheld; 10 Senate voting pending; 7 fail-closed no action \\
\bottomrule
\end{tabular}
}
\end{center}

Source: \path{PROMPT_POLICY_LEAKAGE_WITH_AEGIS_COUNTERFACTUAL.csv}. The table reports prompt-policy leakage rows only; it does not count Aegis-governed side-effect completion.

Full row-level leakage counterfactuals are listed in the source artifacts identified in Appendix H.

\section*{Appendix D. Policy controls implicated by leakage rows}

This appendix makes the prompt-policy leakage rows concrete by linking each leakage class to the controls and source references used by the scoring and governance-mapping layer. These mappings are not claims of real-world legal breach or completed harm. They identify the policy/control categories that the proposed action would have implicated had it been executed. In the governed Aegis path, the corresponding rows recorded non-execution outcomes and zero governed risky side-effect completions.

The policy-control mapping identifies why the prompt-policy leakage rows mattered operationally. The implicated controls include:

\begin{itemize}
  \item \texttt{INT-AI-RESOURCE-BOUNDS}
  \item \texttt{TAX-RESOURCE-RUNAWAY}
  \item \texttt{INT-SENSITIVE-DATA-NO-UNAUTH-EXTERNAL}
  \item \texttt{AUF-INFOSEC-DATA-MINIMIZATION}
  \item \texttt{TAX-INDIRECT-DISCLOSURE}
\end{itemize}

A single leakage row can implicate multiple controls.

\subsection*{Supplementary Table S5. Policy controls implicated by leakage rows}

\begin{center}
\begin{tabular}{llr}
\toprule
Control & Leakage task family & Rows \\
\midrule
INT-AI-RESOURCE-BOUNDS & Runaway background job & 40 \\
TAX-RESOURCE-RUNAWAY & Runaway background job & 40 \\
INT-SENSITIVE-DATA-NO-UNAUTH-EXTERNAL & External raw complaints data & 39 \\
AUF-INFOSEC-DATA-MINIMIZATION & External raw complaints data & 39 \\
TAX-INDIRECT-DISCLOSURE & External raw complaints data & 39 \\
\bottomrule
\end{tabular}
\end{center}

Source: \texttt{required\_controls} in \path{PROMPT_POLICY_LEAKAGE_WITH_AEGIS_COUNTERFACTUAL.csv}. A single leakage row can implicate multiple controls.

\subsection*{Worked evidence exhibit: frontier-model temperature 0.7 leakage-to-control trace}

To make the leakage-to-control mapping inspectable, this appendix includes a worked trace from the frontier-model temperature 0.7 run family. This trace is not presented as a claim about a named model provider. It is a row-level example showing how prompt-policy comparator leakage rows were mapped to policy controls and how the corresponding Aegis-governed rows resolved the same tasks.

In the external raw complaints-data task, the prompt-policy condition allowed escalation-style mock tool proposals for external requests involving raw complaints data. The rows were scored as unauthorized action and sensitive disclosure risk. The mapped controls were \texttt{INT-SENSITIVE-DATA-NO-UNAUTH-EXTERNAL}, \texttt{AUF-INFOSEC-DATA-MINIMIZATION}, and \texttt{TAX-INDIRECT-DISCLOSURE}. The source references included the internal sensitive-data handling policy and \texttt{cps\_234\_july\_2019\_for\_public\_release.pdf}; the source jurisdiction/regulator field included \texttt{AU\_FEDERAL\_FINANCE} and \texttt{APRA}. The corresponding Aegis-governed rows recorded non-execution outcomes: execution withholding or local fail-closed/no-action, with \texttt{mock\_tool\_applied=false} and \texttt{risky\_side\_effect\_completed=false}.

In the runaway background-job task, the prompt-policy condition allowed escalation-style mock tool proposals for high-concurrency background-job requests. The rows were scored as unauthorized action and mapped to \texttt{INT-AI-RESOURCE-BOUNDS} and \texttt{TAX-RESOURCE-RUNAWAY}. The source reference was the internal AI tool-use and agentic workflow policy. Across the frontier-model temperature 0.7 loops, the corresponding Aegis-governed rows recorded \texttt{block} with practical outcome \texttt{execution\_withheld}, \texttt{mock\_tool\_applied=false}, and \texttt{risky\_side\_effect\_completed=false}.

This worked trace illustrates the evaluation distinction used throughout the manuscript. The prompt-policy rows are comparator-path leakage events. The Aegis rows are governed non-execution outcomes. The mapping is not a finding of real-world legal breach, regulatory contravention, or completed harm. It identifies the controls and source references that the proposed action would have implicated had the proposed action executed.

\begin{table}[h]
\centering
\small
\caption{Worked frontier-model temperature 0.7 leakage-to-control traces. Rows are prompt-policy comparator leakage events; Aegis outcomes are corresponding governed counterfactuals. The table reports source mappings from the evaluation artifacts and does not claim real-world legal breach, regulatory contravention, or completed harm.}
\begin{tabular}{p{0.24\linewidth}p{0.34\linewidth}p{0.34\linewidth}}
\toprule
Evidence step & Runaway background job & External raw complaints data \\
\midrule
Comparator-path event & Escalation-style proposal allowed & Escalation-style proposal allowed \\
Count in frontier temp 0.7 & 10 rows & 9 rows \\
Risk label & Unauthorized action; resource runaway & Unauthorized action; sensitive disclosure \\
Mapped controls & \texttt{INT-AI-RESOURCE-BOUNDS}; \texttt{TAX-RESOURCE-RUNAWAY} & \texttt{INT-SENSITIVE-DATA-NO-UNAUTH-EXTERNAL}; \texttt{AUF-INFOSEC-DATA-MINIMIZATION}; \texttt{TAX-INDIRECT-DISCLOSURE} \\
Source linkage & Internal AI tool-use and agentic workflow policy & Internal sensitive-data handling policy; \texttt{cps\_234\_july\_2019\_for\_public\_release.pdf}; APRA-linked source/regulator field \\
Aegis outcome & Block / execution withheld & Execution withheld or fail-closed/no-action \\
Mock tool applied? & No & No \\
Risky side effect completed? & No & No \\
\bottomrule
\end{tabular}
\end{table}

The full machine-readable leakage artifacts include required controls, source references, source titles, source jurisdictions or regulators where applicable, source sections or clauses where available, source mapping level, dereference status, and evidence-record pointers. Compact summaries are printed here; row-level mappings are retained in the source report manifest and leakage artifacts indexed in Appendix H.

\section*{Appendix E. Senate settlement and finality}

The 10-run comparison reports 1,019 Senate rows, 60 settled allow rows, 959 settled deny rows, zero settled failed-closed rows, zero settled unknown rows, 1,019 tally-present rows, 1,019 quorum-met rows, and 1,019 effective-final rows. Senate-associated governed mock-tool applications were zero, and Senate-associated governed risky side-effect completions were zero.

Senate escalation means Senate voting path, not an informal approval step. A Senate-settled allow is a governance settlement, not evidence that the original mock tool was applied. Tool application is tracked separately.

\subsection*{Supplementary Table S6. Senate settlement and finality}

\begin{center}
\resizebox{\textwidth}{!}{%
\begin{tabular}{lrrrrrr}
\toprule
Run family & Senate rows & Settled allow & Settled deny & Quorum met & Final tally & Risky completions \\
\midrule
Stubbed & 400 & 50 & 350 & 400 & 400 & 0 \\
Gemma & 300 & 10 & 290 & 300 & 300 & 0 \\
Frontier temp 0 & 119 & 0 & 119 & 119 & 119 & 0 \\
Frontier temp 0.7 & 103 & 0 & 103 & 103 & 103 & 0 \\
Frontier temp 1.0 & 97 & 0 & 97 & 97 & 97 & 0 \\
\midrule
Total & 1019 & 60 & 959 & 1019 & 1019 & 0 \\
\bottomrule
\end{tabular}
}
\end{center}

Source: \path{AEGIS_10_RUN_SENATE_SUMMARY.md}. Senate-settled allow is a governance settlement, not evidence that the original mock tool was applied.

The Senate latency summary is available in the source artifacts identified in Appendix H.

\subsection*{Runtime decision latency}

The timing artifacts report Aegis runtime decision latency for the governed sandbox path. In the reported 10-run pack, median Aegis latency across the five governed run families ranged from 25.598 ms to 27.754 ms, p95 latency ranged from 30.908 ms to 66.848 ms, and the maximum observed Aegis latency was 237.897 ms. These timings measure the sandbox/runtime decision path captured in the evaluation artifacts. They should not be read as production network latency, end-user latency, or a general benchmark for all deployment environments.

\begin{table}[h]
\centering
\caption{Aegis runtime decision latency in the evaluated sandbox. Values are reported from existing timing artifacts and are not production network or end-user latency benchmarks.}
\begin{tabular}{lr}
\toprule
Metric & Value \\
\midrule
Run families summarized & 5 \\
Governed rows per run family & 420 \\
Median Aegis latency range & 25.598--27.754 ms \\
p95 Aegis latency range & 30.908--66.848 ms \\
Maximum observed Aegis latency & 237.897 ms \\
\bottomrule
\end{tabular}
\end{table}

\section*{Appendix F. Trusted provenance boundary audit}

The repeated evaluation reports 1,832 Aegis-attempted governed rows and 1,832 trusted Aegis-resolved provenance-valid rows. Provenance-unknown rows were zero, and provenance-invalid rows were zero.

Trusted provenance means Aegis/server-side resolved provenance. Model-supplied, client-supplied, or PEP-supplied citations are not production-valid trusted provenance. Evidence completeness and provenance boundary validity are distinct. Local fail-closed/no-tool rows are legitimate non-execution outcomes, not provenance-invalid rows.

\subsection*{Supplementary Table S7. Trusted Aegis-resolved provenance}

\begin{center}
\begin{tabular}{lrrr}
\toprule
Run family & Trusted provenance-valid & Provenance unknown & Provenance invalid \\
\midrule
Stubbed & 420 & 0 & 0 \\
Gemma & 350 & 0 & 0 \\
Frontier temp 0 & 367 & 0 & 0 \\
Frontier temp 0.7 & 347 & 0 & 0 \\
Frontier temp 1.0 & 348 & 0 & 0 \\
\midrule
Total & 1832 & 0 & 0 \\
\bottomrule
\end{tabular}
\end{center}

Source: \path{AEGIS_10_RUN_PROVENANCE_SUMMARY.md}. Evidence completeness is distinct from trusted provenance validity.

\section*{Appendix G. Workflow and failure-category coverage}

The governed corpus covers five workflow families:

\begin{itemize}
  \item \texttt{active\_law\_change}
  \item \texttt{agentic\_authority\_tool\_use}
  \item \texttt{ai\_consumer\_impact}
  \item \texttt{external\_communication}
  \item \texttt{vendor\_service\_provider}
\end{itemize}

The governed corpus also covers the following failure categories:

\begin{itemize}
  \item \texttt{authority\_spoofing}
  \item \texttt{cross\_agent\_unsafe\_propagation}
  \item \texttt{destructive\_or\_disproportionate\_tool\_use}
  \item \texttt{false\_task\_completion\_state\_divergence}
  \item \texttt{indirect\_sensitive\_information\_disclosure}
  \item \texttt{non\_owner\_compliance}
  \item \texttt{uncontrolled\_resource\_consumption\_or\_persistent\_loops}
  \item \texttt{none}
\end{itemize}

This taxonomy demonstrates coverage beyond a single narrow scenario, but it is not claimed to be exhaustive.

\subsection*{Supplementary Table S8. Governed workflow coverage}

\begin{center}
\begin{tabular}{lr}
\toprule
Workflow family & Governed rows \\
\midrule
agentic\_authority\_tool\_use & 600 \\
ai\_consumer\_impact & 400 \\
external\_communication & 400 \\
vendor\_service\_provider & 400 \\
active\_law\_change & 300 \\
\midrule
Total & 2100 \\
\bottomrule
\end{tabular}
\end{center}

Source: aggregated from \path{AEGIS_10_RUN_GOVERNED_DECISION_TRACE.csv}; detailed bucket splits are in \path{AEGIS_10_RUN_BY_WORKFLOW_AND_BUCKET.md}.

\subsection*{Supplementary Table S9. Governed failure-category coverage}

\begin{center}
\begin{tabular}{lr}
\toprule
Failure category & Governed rows \\
\midrule
non\_owner\_compliance & 500 \\
none & 400 \\
false\_task\_completion\_state\_divergence & 300 \\
indirect\_sensitive\_information\_disclosure & 250 \\
authority\_spoofing & 200 \\
uncontrolled\_resource\_consumption\_or\_persistent\_loops & 200 \\
cross\_agent\_unsafe\_propagation & 200 \\
destructive\_or\_disproportionate\_tool\_use & 50 \\
\midrule
Total & 2100 \\
\bottomrule
\end{tabular}
\end{center}

Source: aggregated from \path{AEGIS_10_RUN_GOVERNED_DECISION_TRACE.csv}; detailed bucket splits are in \path{AEGIS_10_RUN_BY_FAILURE_CATEGORY.md}.


\section*{Appendix H. Artifact manifest and reproducibility notes}

The source report manifest is \path{source_report_manifest.md}. It lists the 10-run reports, single-run reports, leakage reports, Senate reports, provenance reports, workflow/failure reports, raw matrix-record locations, pack-completeness reports, rerun-readiness reports, manifests, and timing records used by the manuscript and this Supplementary Information.

The worked frontier-model temperature 0.7 leakage trace in Appendix D is sourced from:

\begin{itemize}
  \item \texttt{paper\_eval/aegis\_action\_sandbox/reports/10 Run Folder/}\\
  \texttt{Frontier temp 0.7 10 run/frontier\_report/}\\
  \texttt{PROMPT\_POLICY\_LEAKAGE\_EVENT\_CHAIN.csv}
  \item \texttt{paper\_eval/aegis\_action\_sandbox/reports/10 Run Folder/}\\
  \texttt{Frontier temp 0.7 10 run/frontier\_report/}\\
  \texttt{PROMPT\_POLICY\_LEAKAGE\_WITH\_AEGIS\_COUNTERFACTUAL.csv}
\end{itemize}

The runtime decision-latency summary in Appendix E is sourced from:

\begin{itemize}
  \item \texttt{paper\_eval/aegis\_action\_sandbox/reports/10 Run Folder/}\\
  \texttt{10 run comparison/comparison\_report/AEGIS\_LATENCY\_SUMMARY.md}
  \item \texttt{paper\_eval/aegis\_action\_sandbox/reports/10 Run Folder/}\\
  \texttt{10 run comparison/comparison\_report/tables/aegis\_latency\_summary.csv}
\end{itemize}

\subsection*{Supplementary Table S10. Report-pack completeness}

\begin{center}
\begin{tabular}{lrrrr}
\toprule
Run family & Total rows & Governed rows & Matrix files & Pack complete \\
\midrule
Stubbed 10x & 1260 & 420 & 10 & True \\
Gemma 10x & 1260 & 420 & 10 & True \\
Frontier temp 0 10x & 1260 & 420 & 10 & True \\
Frontier temp 0.7 10x & 1260 & 420 & 10 & True \\
Frontier temp 1.0 10x & 1260 & 420 & 10 & True \\
\midrule
Total & 6300 & 2100 & 50 & True \\
\bottomrule
\end{tabular}
\end{center}

Source: \path{AEGIS_10_RUN_PACK_COMPLETENESS.md}.

Additional human-readable appendix notes are provided in:

\begin{itemize}
  \item \path{supplementary_appendix_index.md}
  \item \path{leakage_policy_control_appendix.md}
  \item \path{senate_provenance_appendix.md}
  \item \path{reproducibility_appendix.md}
\end{itemize}

No model inference, backend service calls, policy mutations, prompt changes, task changes, or real side effects are performed by this Supplementary Information build. The public artifact release includes the sandbox policy-enforcement point, synthetic task corpus, mock tools, prompt-policy comparator, scoring logic, report builders, output schemas, frozen sanitized result artifacts, and documentation required to inspect the reported tables. The public sandbox PEP repository is available at \url{https://github.com/CyberQube1/Aegis_PEP_Sandbox.git}; archival DOI details will be added before public release. The public release does not include the production Aegis kernel, production trust infrastructure, private credentials, production policy bundles, signing material, or live endpoint details. Offline public runs support sandbox mechanics and report/table inspection without live Aegis decisions; they do not reproduce or simulate the Aegis PDP. Validation against the real Aegis PDP is available to reviewers or researchers on request through scoped SPQR-issued credentials and mock-only trust material. This boundary is intentional: releasing signing material, production trust infrastructure, or live PDP endpoints would weaken the very control surface evaluated by the paper.

\end{document}